%% file: main.tex
\documentclass[sigconf]{acmart}
\usepackage{subcaption}
\usepackage{booktabs}
\AtBeginDocument{%
  }

\setcopyright{acmlicensed}
\copyrightyear{2025}
\acmYear{2025}
\setcopyright{cc}
\setcctype{by}
\acmConference[KDD '25]{Proceedings of the 31st ACM SIGKDD Conference on Knowledge Discovery and Data Mining V.2}{August 3--7, 2025}{Toronto, ON, Canada}
\acmBooktitle{Proceedings of the 31st ACM SIGKDD Conference on Knowledge Discovery and Data Mining V.2 (KDD '25), August 3--7, 2025, Toronto, ON, Canada}
\acmDOI{10.1145/3711896.3736822}
\acmISBN{979-8-4007-1454-2/2025/08}
\begin{document}

\title{A Scalable Pretraining Framework for Link Prediction with Efficient Adaptation}

\author{Yu Song}
\affiliation{%
\department{Computer Science and Engineering}
  \institution{Michigan State University}
  \city{East Lansing}
  \state{MI}
  \country{USA}}
\email{songyu5@msu.edu}

\author{Zhigang Hua}
\affiliation{%
\department{Monetization AI}
  \institution{Meta}
  \city{Sunnyvale}
  \state{CA}
  \country{USA}}
\email{zhua@meta.com}

\author{Harry Shomer}
\affiliation{%
 \department{Computer Science and Engineering}
  \institution{Michigan State University}
  \city{East Lansing}
  \state{MI}
  \country{USA}}
\email{shomerha@msu.edu}

\author{Yan Xie}
\affiliation{%
\department{Monetization AI}
  \institution{Meta}
  \city{Sunnyvale}
  \state{CA}
  \country{USA}}
\email{yanxie@meta.com}

\author{Jingzhe Liu}
\affiliation{%
 \department{Computer Science and Engineering}
  \institution{Michigan State University}
  \city{East Lansing}
  \state{MI}
  \country{USA}}
\email{liujin33@msu.edu}

\author{Bo Long}
\affiliation{%
  \department{Monetization AI}
  \institution{Meta}
  \city{Menlo Park}
  \state{CA}
  \country{USA}}
\email{bolong@meta.com}

\author{Hui Liu}
\affiliation{%
  \department{Computer Science and Engineering}
  \institution{Michigan State University}
  \city{East Lansing}
  \state{MI}
  \country{USA}}
\email{liuhui7@msu.edu}

\renewcommand{\shortauthors}{Yu Song et al.}

\begin{abstract}
\input{sections/abstract}
\end{abstract}




\maketitle

\newcommand\kddavailabilityurl{https://doi.org/10.5281/zenodo.15597907}

\ifdefempty{\kddavailabilityurl}{}{
\begingroup\small\noindent\raggedright\textbf{KDD Availability Link:}\\
The source code of this paper has been made publicly available at \url{\kddavailabilityurl}.
\endgroup
}

\section{Introduction}
\input{sections/intro}

\section{Related Work}
\input{sections/related_work_main}

\section{Preliminaries}
\input{sections/preliminary}

\input{sections/prelim_study}

\section{Method}
\input{sections/method}

\section{Experiments}
\input{sections/experiments}

\section{Conclusion}
\input{sections/conclusion}

\section*{Acknowledgements}
Yu Song, Harry Shomer, Jingzhe Liu, and Hui Liu are supported by the National Science Foundation (NSF) under grant numbers CNS2321416, IIS2212032, IIS2212144, IOS2107215, 
DUE2234015, 

\noindent CNS2246050, DRL2405483 and IOS2035472, US Department of Commerce, Gates Foundation, the Michigan Department of Agriculture and Rural Development, Amazon, Meta and SNAP.

\bibliographystyle{ACM-Reference-Format}
\balance
\bibliography{ref}

\appendix
\include{sections/appendix}

\end{document}

%% file: sections/abstract.tex
Link Prediction (LP) is a critical task in graph machine learning. While Graph Neural Networks (GNNs) have significantly advanced LP performance recently, existing methods face key challenges including limited supervision from sparse connectivity, sensitivity to initialization, and poor generalization under distribution shifts.  

We explore pretraining as a solution to address these challenges. 
Unlike node classification, LP is inherently a pairwise task, which requires the integration of both node- and edge-level information. 
In this work, we present the first systematic study on the transferability of these distinct modules and propose a late fusion strategy to effectively combine their outputs for improved performance. To handle the diversity of pretraining data and avoid negative transfer, we introduce a Mixture-of-Experts (MoE) framework that captures distinct patterns in separate experts, facilitating seamless application of the pretrained model on diverse downstream datasets. For fast adaptation, we develop a parameter-efficient tuning strategy that allows the pretrained model to adapt to unseen datasets with minimal computational overhead. Experiments on 16 datasets across two domains demonstrate the effectiveness of our approach, achieving state-of-the-art performance on low-resource link prediction while obtaining competitive results compared to end-to-end trained methods, with over 10,000x lower computational overhead.

%% file: sections/intro.tex
Link Prediction (LP) is a fundamental task in graph learning with broad applications across social networks, biology, recommendation systems, and beyond~\cite{kovacs2019network, daud2020applications, huang2005link,ma2021deep}. Traditional LP methods rely on hand-crafted heuristics to model pairwise node relationships, such as common neighbors or shortest path distances~\cite{newman2001clustering, katz1953new}. More recently, these heuristics have been integrated with node representations learned by Graph Neural Networks (GNNs), giving rise to the GNN4LP paradigm~\cite{chamberlain2022graph, wang2023neural, yun2021neo, shomer2024lpformer}. While these methods have achieved strong performance, they still face several key challenges, such as limited supervision due to sparse graph connectivity, sensitivity to model initialization and hyperparameter choices, and poor generalization under distribution shifts~\cite{li2024evaluating}.

Pretraining offers a promising solution to address these limitations. 
By learning generalized patterns from large-scale data, pretrained models provide well-initialized parameters that can be effectively adapted to unseen tasks with minimal fine-tuning or zero-shot learning, with a record of success in Computer Vision (CV) \cite{CLIP, kirillov2023segment, radford2021learning} and Natural Language Processing (NLP) \cite{Brown2020LanguageMA, devlin2018bert}.
In the graph domain, there has been growing interest in developing foundation models capable of generalizing across diverse datasets and tasks~\cite{liu2023one, huang2024prodigy, li2024zerog, chen2024llaga, song2024pure, xia2024anygraph}. However, the majority of existing graph pretraining efforts focus on node classification, with few approaches tailored for link prediction~\cite{dong2024universal, he2024linkgpt}.

Unlike node classification which primarily relies on node-level representations to make predictions, LP is inherently a pairwise task where the formation of links depends on the interactions between nodes (i.e.,``pairwise information'')~\cite{shomer2024lpformer, mao2023revisiting}. To capture both node- and edge-level signals, existing GNN4LP methods typically employ two complementary modules: one to learn node representations and another to encode edge features, which are then fused via a shared score function to estimate the probability of edge existence. 
While this design has proven effective in standard supervised settings, the respective roles and contributions of these components during pretraining remain unclear. This gives rise to the first key challenge in LP pretraining: {\bf (1)} {\it How do different modules contribute to the pretraining process, and how can they be combined to maximize performance?}

Pretraining often involves large-scale data with diverse distributions. While scaling laws in CV and NLP suggest consistent improvements with increased data~\cite{kaplan2020scaling, abnar2022exploring, zhai2022scaling}, graph pretraining exhibits more nuanced behaviors. Prior work has shown that simply adding more graphs does not always yield better downstream performance and can even result in negative transfer, especially when distribution shifts are present between pretraining and downstream tasks~\cite{cao2023pre, xu2023betterWITHLESS}.
This introduces the second challenge: {\bf (2)} {\it How can we flexibly absorb diverse knowledge from large-scale pretraining data while ensuring compatibility with downstream datasets?} 

Adapting a pretrained model to a new dataset also presents its difficulties~\cite{fang2025benefits, li2025unveiling}. 
Full Fine-tuning can be computationally expensive and prone to catastrophic forgetting and overfitting~\cite{kirkpatrick2017overcoming}. In contrast, zero-shot learning lacks the flexibility to to capture dataset-specific nuances, resulting in suboptimal performance~\cite{xie2025one}.
This raises the third challenge: {\bf (3)} \textit{how to efficiently adapt a pretrained LP model to new graphs while preserving its learned knowledge?}

To address the first challenge, we conduct a preliminary study to evaluate the transferability of pretrained LP models across diverse downstream datasets (see Section~\ref{subsec:trans}).
Our empirical results show that by pretraining on a large-scale dataset, both the node and edge modules generalize well to unseen graphs. 
To improve module integration, we identify an imbalanced training issue and propose a late fusion strategy for enhanced performance.
To promote transferability, we propose a Mixture-of-Experts (MoE) framework, where each expert captures distinct patterns from the pretraining data, allowing the model to harness diverse knowledge while mitigating conflicts and negative transfer. For adaptation, we develop a parameter-efficient tuning strategy that learns only the expert assignment for each downstream dataset while keeping the expert parameters unchanged, enabling adaptive expert selection with minimal computational overhead.  
Our contributions can be summarized as follows:
\begin{itemize}
\item We present the first systematic study of pretraining specifically for link prediction, analyzing the transferability of node and edge modules and proposing effective fusion strategies.
\item We introduce a novel Mixture-of-Experts (MoE) framework for LP pretraining, along with a parameter-efficient adaptation method that enables flexible transfer to downstream datasets.
\item We validate our approach through extensive experiments on 16 datasets spanning two domains, demonstrating its effectiveness and generalizability.
\end{itemize}

%% file: sections/related_work_main.tex
\textbf{Link Prediction.}  
Traditional link prediction methods rely on heuristic metrics that extract structural features from graph topology, such as Katz index and clustering coefficients~\cite{katz1953new, newman2001clustering, li2020distance}. More recent approaches improve upon these by incorporating Graph Neural Networks (GNNs), either through edge-personalized message passing~\cite{zhang2018link, zhu2021neural} or by modeling pairwise interactions alongside node representations~\cite{yun2021neo, wang2023neural, Wang2024SubgraphPT}. However, these methods follow a \textit{"one model, one dataset"} paradigm, where a separate model must be trained for each dataset, resulting in poor cross-graph transferability. To the best of our knowledge, we are the first to explore transfer learning for general link prediction, where both node- and edge-level information are transferred via learnable models. Prior efforts in this space are limited: \cite{dong2024universal} investigates in-context learning for LP using structural features but ignores node attributes, while \cite{he2024linkgpt} employs large language models for LP on text-attributed graphs, which is computationally expensive and lacks efficient pairwise feature utilization.  

\noindent\textbf{Graph Foundation Models.}  
Graph Foundation Models (GFMs) aim to unify graph representation learning across diverse datasets and tasks \cite{liu2023one, huang2024prodigy, li2024zerog, chen2024llaga, song2024pure, xia2024anygraph}. While these methods share the philosophy of \textit{"pretrain once, serve all"}, they primarily focus on generating node-level representations, which are inherently suboptimal for link prediction~\cite{zhang2021labeling}. In addition, subgraph-based approaches~\cite{liu2023one, li2024zerog, chen2024llaga} incur high computational costs due to the need for subgraph extraction for each query, while models such as~\cite{li2024zerog} require fine-tuning large language models, making them computationally expensive and less scalable.
In contrast, our work fills this gap by developing a scalable and transferable framework that jointly captures node- and edge-level dependencies, representing a significant step toward extending GFMs to pairwise prediction tasks.

\begin{figure*}[t]
    \centering
    \begin{subfigure}[b]{0.3\textwidth}
        \centering
        \resizebox{\textwidth}{3.2cm}{\includegraphics{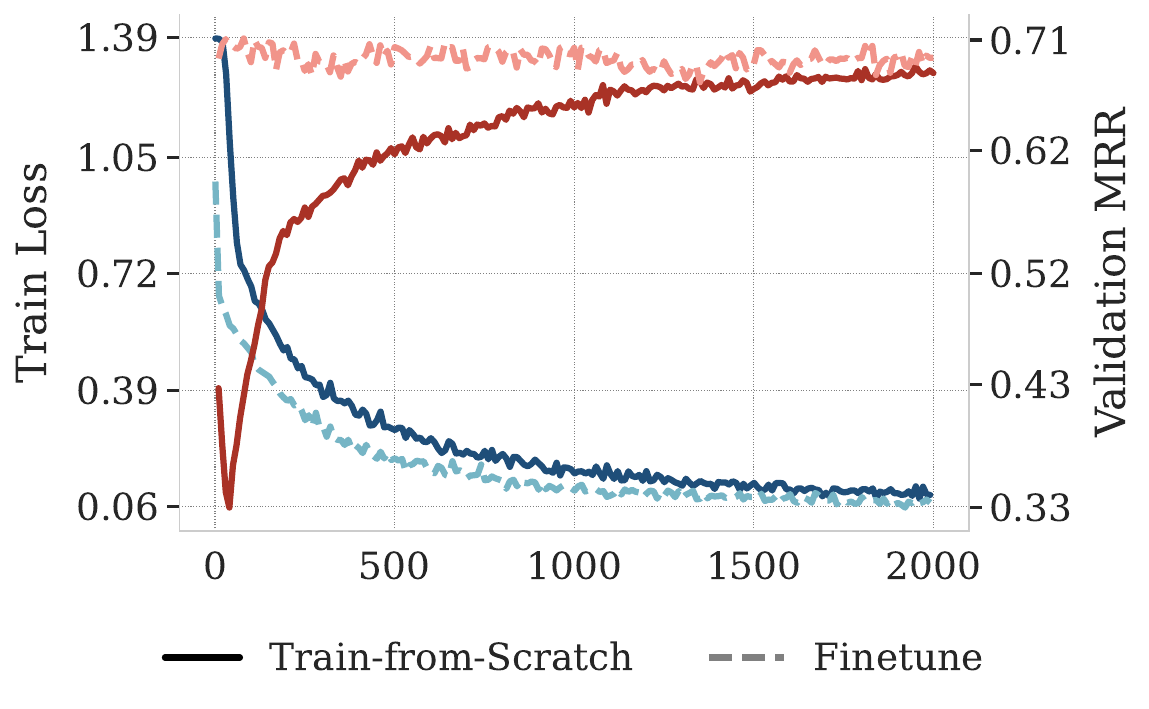}}  
        \caption{Training dynamics}
        \label{fig:pretrain_dynamics}
    \end{subfigure}
    \hfill
    \begin{subfigure}[b]{0.69\textwidth}
        \centering
        \resizebox{1.0\textwidth}{3.2cm}{\includegraphics{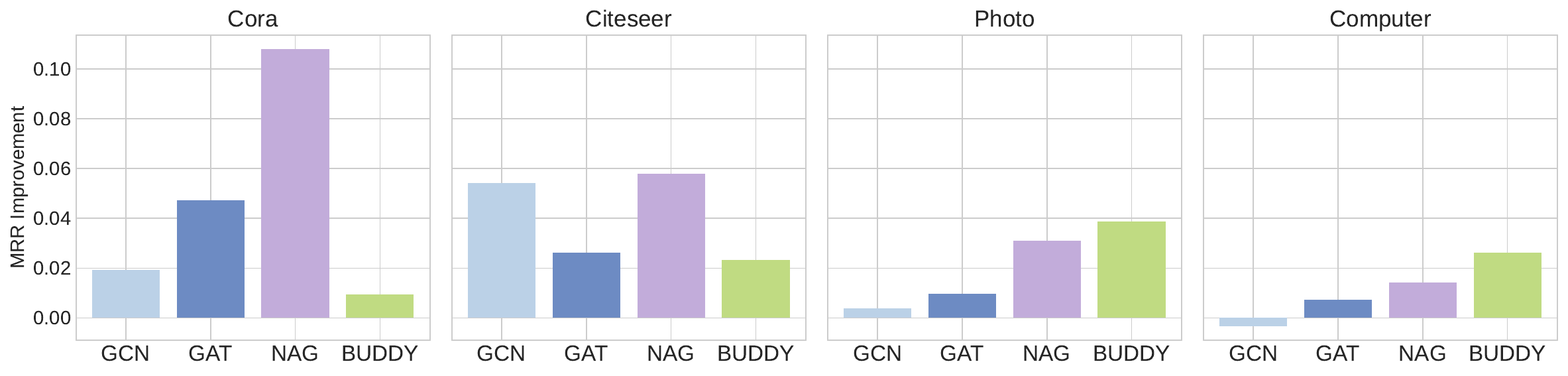}}  
        \caption{Improvements of pretraining}
        \label{fig:pretrain_improvements}
    \end{subfigure}
    \caption{Pretraining improves link prediction. (a) Training dynamics of a two-layer GCN with and without pretraining on the Cora dataset. The finetuned model achieves better performance and faster convergence.  (b) Improvements of pretrain-then-finetune over training from scratch, measured by test MRR after convergence. Pretrained checkpoints boost final performance with different models and datasets. }
    \label{fig:pretrain}
\end{figure*}

%% file: sections/preliminary.tex
 \subsection{Background}
Link prediction aims to predict missing links between node pairs in partially observed graphs. Given a graph $G$, an adjacency matrix of the observed edges $A \in \mathbb{R}^{n \times n}$, and a feature matrix $X \in \mathbb{R}^{n \times d}$ describing the features of nodes, LP predicts the probability of forming an edge $(i, j)$. Previous study shows that links form due to various underlying mechanisms, which can be largely categorized into \textbf{feature proximity (FP)} and \textbf{structure proximity (SP)} \cite{mao2023revisiting}.
FP corresponds to the feature similarity between nodes, reflecting the homophily assumption that nodes with similar characteristics are more likely to connect. FP can be effectively captured by powerful node encoders like Message Passing Neural Networks (MPNNS)~\cite{gilmer2017neural} to produce high-quality node embeddings, along with a score function that computes the probability of an edge existing between a pair of nodes. Formally, this process can be written as: 
\begin{equation}
H = \text{NodeEncoder}(A, X), \quad p_{ij} = \text{ScoreFunction}(H_i \odot H_j)
\end{equation}
\noindent where $p_{ij}$ is the probability of the link existing and $\odot$ denotes Hadamard product. However, studies on MPNN expressiveness indicate that standard MPNNs are unable to count triangles, which in turn limits their ability to compute common neighbors and other heuristics such as Adamic-Adar (AA) and Resource Allocation (RA), which are key predictors of link formation\cite{srinivasan2019equivalence, zhang2021labeling}. Pairwise encodings, on the other hand, have proven effective to represent structure proximity, capturing the local or global relationships between pairs of nodes through overlap of neighbors or path information \cite{katz1953new, newman2001clustering, li2020distance}. Link prediction with pairwise encodings can be expressed as:
\begin{equation}
e_{ij} = \text{EdgeEncoder}(A, i, j), \quad p_{ij} = \text{ScoreFunction}(e_{ij})
\end{equation}
To complement information from both sources, recent efforts have attempted to fuse the signals from both \cite{Wang2024SubgraphPT, yun2021neo, wang2023neural, shomer2024lpformer}. This is done by integrating the outputs of the node and edge modules before feeding them to a shared score function:
\begin{equation}
    p_{ij} = \text{ScoreFunction}(H_i \odot H_j \mid e_{ij}).
    \label{eq:early_fusion}
\end{equation}

Before developing a pretraining strategy for LP models, it is essential to first understand the capabilities and limitations of existing frameworks. In particular, we need to assess the transferability of node and edge modules, as well as the effectiveness of different fusion strategies in combining feature and structural information. Understanding these components is key to identifying the principles that underpin effective and robust LP pretraining. In the next section, we conduct a systematic empirical analysis to inform the design of our pretraining framework.







%% file: sections/prelim_study.tex
\subsection{Transferability Study}
\label{subsec:trans}


Unlike other graph tasks, LP models involve both a node module and an edge module. In this section, we investigate the transferability of each component \textit{independently}, isolating their contributions during pretraining. Specifically, we pretrain the two modules separately and fine-tune the obtained checkpoints on various downstream datasets. For the node module, we choose GCN \cite{kipf2016semi}, GAT \cite{velivckovic2017graph} and a state-of-the-art graph transformer---NAGphormer \cite{chen2022nagphormer}. For the edge module, we employ the non-learnable structural encoding from BUDDY \cite{Wang2024SubgraphPT} as edge features, due to its capability of capturing a diverse set of LP heuristics.
These models are selected based on their strong empirical performance and computational efficiency, both of which are essential for large-scale pretraining.
For all methods, we adopt a 3-layer MLP as the score function to compute the final edge probability.
We pretrain all models on \textsf{ogbn-papers100M} \cite{hu2020openOGB} containing approximately 100M academic publications and their citation relationships, constituting the largest publicly available graph dataset. The pretrained models are then fine-tuned and evaluated on four downstream datasets: Cora, Citeseer, Photo and Computers, where Cora and Citeseer also belong to citation networks while Photo and Computers are extracted from Amazon e-commerce datasets. To provide a unified input space for different graph datasets, we process the original texts with SentenceBERT \cite{reimers2019sentence} and use the obtained textual embeddings as node features, following \cite{Chen2023ExploringTP}. 

Figure \ref{fig:pretrain_dynamics} compares the training dynamics of a two-layer GCN fine-tuned from a pretrained checkpoint versus trained from scratch. The pretrained model provides a significantly better starting point, exhibiting lower training loss and higher validation scores in Mean Reciprocal Rank (MRR).  To further assess the impact of pretraining, we extend the analysis to additional backbone architectures and report their performance gains relative to their train-from-scratch counterparts in Figure~\ref{fig:pretrain_improvements}. Across most settings, pretraining yields consistent improvements, though the extent varies depending on the model architecture and dataset.
Overall, our study on module transferability yields three main insights: (a) LP knowledge can be effectively transferred between datasets ; (b) the transferability is general and agnostic to model architecture, and (c) the transferred knowledge generalizes well across graphs from different domains.

\subsection{Fusion Strategies}
\label{subsec:fusion}

\begin{figure*}[t]
  \centering
  \setlength{\abovecaptionskip}{5pt} 
  \setlength{\belowcaptionskip}{-2pt} 
  \vspace{-3pt} 

  \begin{subfigure}[b]{0.26\textwidth} 
    \centering
    \includegraphics[width=0.9\textwidth]{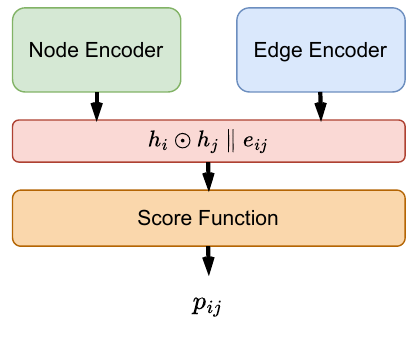} 
    \caption{Early fusion}
    \label{fig:fusion_a}
  \end{subfigure}
  \hfill
  \begin{subfigure}[b]{0.26\textwidth} 
    \centering
    \includegraphics[width=0.9\textwidth]{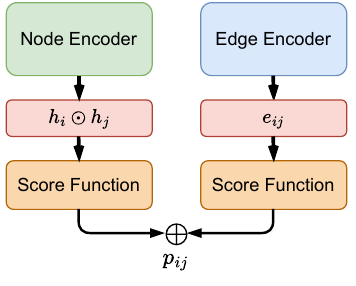} 
    \caption{Late fusion}
    \label{fig:fusion_b}
  \end{subfigure}
  \hfill
  \begin{subfigure}[b]{0.40\textwidth} 
    \centering
    \includegraphics[width=0.9\textwidth]{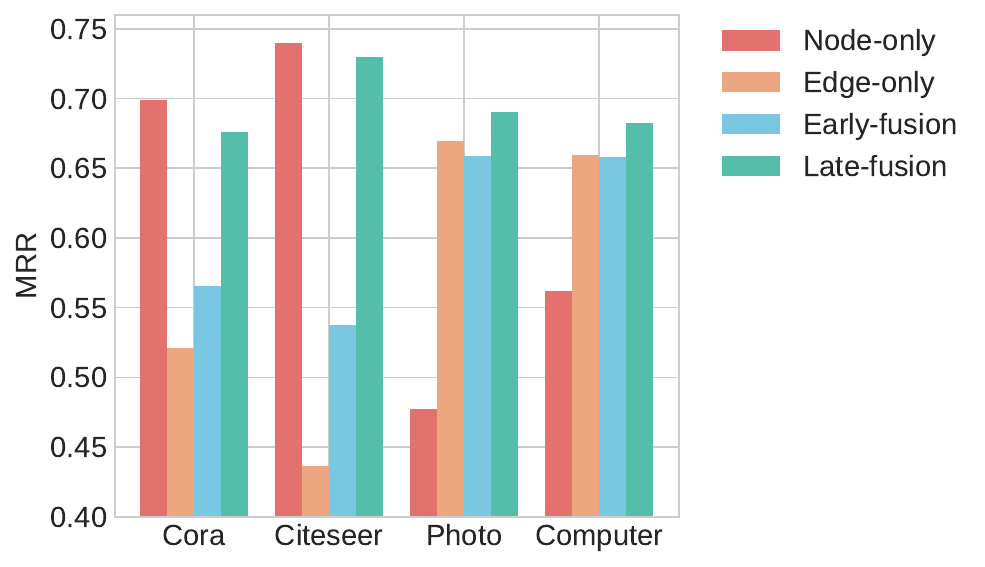} 
    \caption{Comparison of fusion strategies}
    \label{fig:fusion_c}
  \end{subfigure}

  \caption{Comparison of early fusion and late fusion. (a) Early fusion occurs in the representation space and trains the node/edge encoder in an end-to-end manner. (b) Late fusion merges the logits from two independently trained branches. (c) Early fusion lags behind node/edge-only modules, while late fusion improves performance on Photo and Computer.}
  \label{fig:three_subplots_one_row}
  \vspace{-8pt} 
\end{figure*}

\begin{figure}[t]
    \centering
    \begin{subfigure}[b]{0.48\columnwidth}
        \centering
        \includegraphics[width=\linewidth]{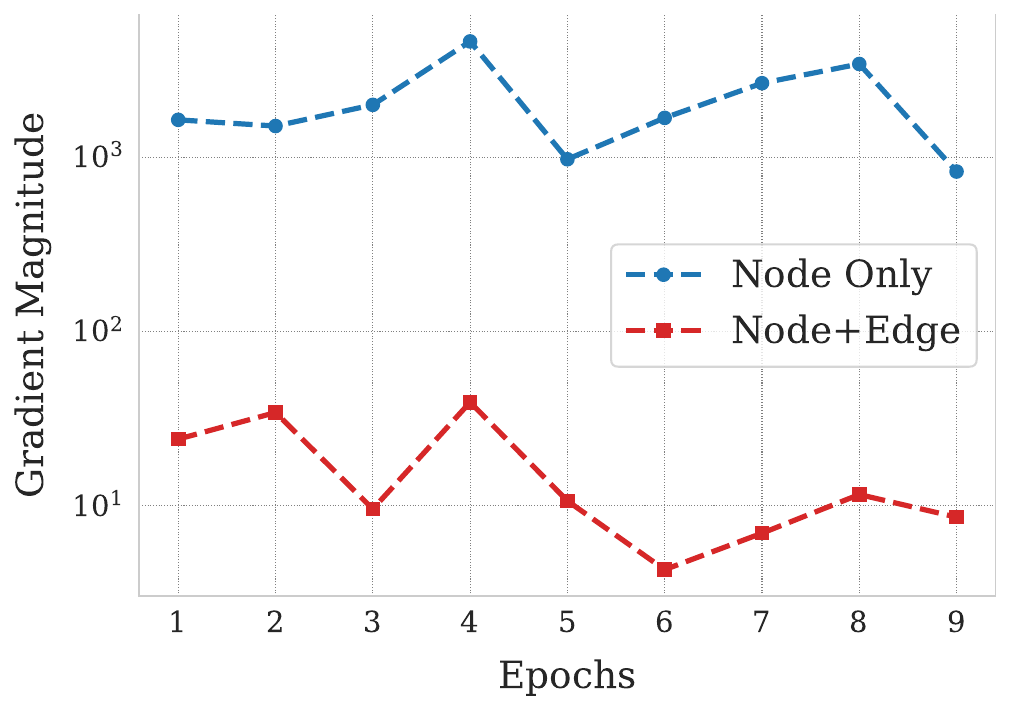}
        \caption{Gradient scales}
        \label{fig:train_a}
    \end{subfigure}
    \hfill
    \begin{subfigure}[b]{0.48\columnwidth}
        \centering
        \includegraphics[width=\linewidth]{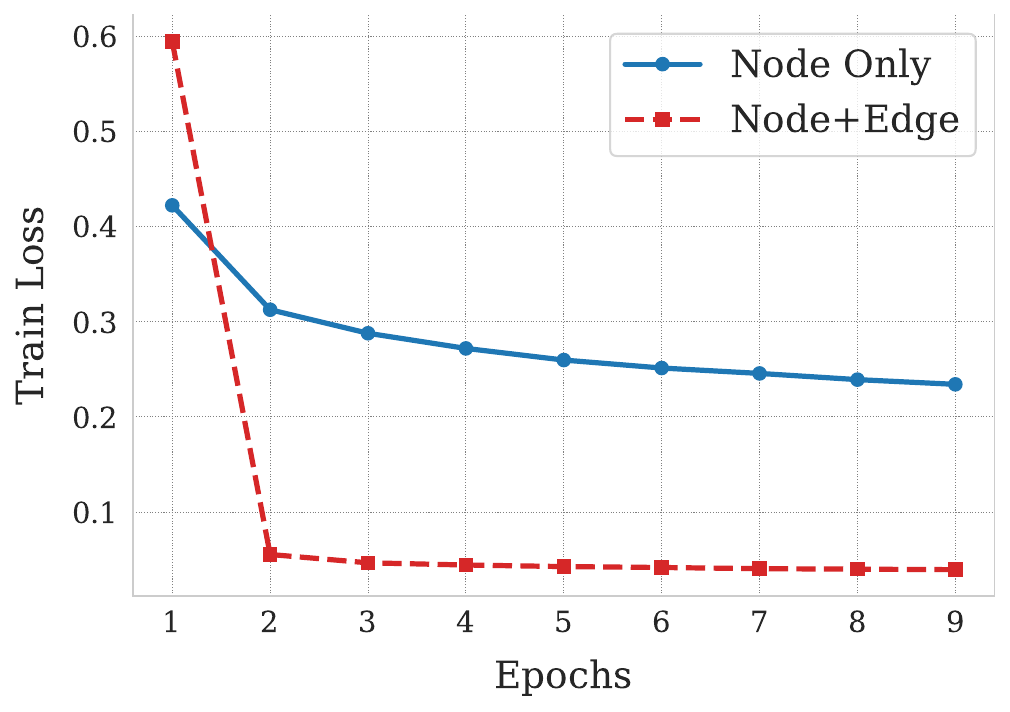}
        \caption{Loss values}
        \label{fig:train_b}
    \end{subfigure}
    \caption{
    Training dynamics of the node-only architecture vs. early fusion (node+edge).
    Early fusion leads to degraded performance due to gradient imbalance.
    }
    \label{fig:training}
\vspace{-5mm}
\end{figure}

In the previous subsection, we demonstrated the transferability of the node and edge modules for LP. A simple strategy would then be to pretrain existing LP methods, \textit{without modification to their framework}. This takes the form of Eq.~\eqref{eq:early_fusion}, where existing GNN4LP methods combine the edge- and node-level representations and pass them to a single score function, a strategy known as \textit{early fusion} ~\cite{shomer2024lpformer, wang2023neural, chamberlain2022graph}. However, its effectiveness in the context of pretraining remains unclear.

In this section, we study two different fusion strategies: \textit{early fusion} and \textit{late fusion}. Early fusion occurs in the embedding space, where node representations and pairwise encodings are concatenated before being processed by a shared score function. Late fusion, on the other hand, aggregates outputs from two independently trained modules, serving as a way of model ensemble. An illustration of these two approaches is presented in Figure \ref{fig:fusion_a} and \ref{fig:fusion_b}.
Following our prior setup, we perform pretraining on \textsf{ogbn-papers100M} and evaluate zero-shot LP performance under different fusion strategies in Figure \ref{fig:fusion_c}.

Surprisingly, early fusion does not improve performance over node-only or edge-only architectures. In fact, it consistently degrades performance across all downstream datasets. 
To better understand this phenomenon, we monitor the training dynamics of early fusion to determine whether each module acquires knowledge as expected during pretraining. In Figure \ref{fig:train_a}, we compare the magnitudes of gradients received by the node encoder under node-only training versus early fusion. We observe that in early fusion, the node encoder receives significantly weaker gradients, which seriously hinders its ability to learn meaningful representations. This occurs because structural features provided by the edge module are highly predictive of link existence, creating an easy pathway for the model to classify edges. As a result, the loss decreases rapidly in early training stages (see Figure \ref{fig:train_b}), leading to insufficient optimization of the node module. This imbalance is a well-known issue in multimodal fusion, which often requires sophisticated techniques to achieve balanced training~\cite{peng2022balanced}.

 We then investigate \textit{late fusion} as a simple yet effective remedy. Unlike early fusion, late fusion aggregates the outputs from the node and edge modules, which are trained \textit{independently} on the same pretraining dataset. 
In our preliminary study, we adopt a simple sum-pooling on the sigmoid-normalized probabilities from both modules. From the results in Figure \ref{fig:fusion_c}, late fusion achieves significantly better performance than early fusion. Notably, it improves the performance of individual modules in certain cases, reflected by the two e-commerce datasets, Photo and Computers. Given this result, we hypothesize that \textit{late fusion can better combine the two types of information}, as it bypasses the optimization pitfalls of early fusion and allows both modules to learn effectively.
However, late fusion also faces challenges. On Cora and Citeseer, it slightly underperforms compared to the node-only baseline, due to its inability to dynamically prioritize between the two modules. In Section \ref{subsec:adaptation}, we address this issue by proposing an adaptive fusion mechanism, which ensures performance never worse than the best individual module, while improving the overall performance when possible.

%% file: sections/method.tex
In this section, we introduce our proposed framework, Pretraining and Adaptation for Link Prediction (PALP), and then details its key components.

\subsection{Overview}

\begin{figure*}[t]
  \centering
  \setlength{\abovecaptionskip}{2pt}  
  \setlength{\belowcaptionskip}{-2pt} 
  \vspace{-3.5pt} 
  \includegraphics[width=0.90\textwidth]{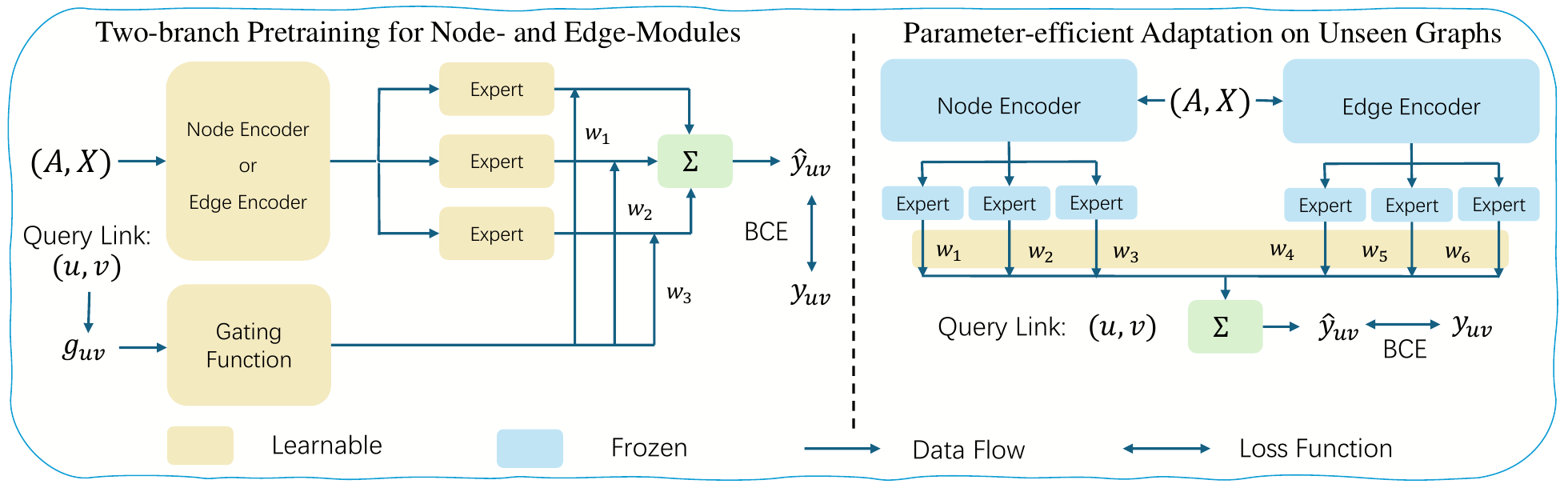} 
  \caption{
Overview of PALP. (Left) During pretraining, the node and edge modules are trained independently. Each module includes an encoder, a gating function to assign edges to experts, and a set of expert score functions. 
(Right) During adaptation, all pretrained components are frozen, and a lightweight weight vector is learned to adaptively aggregate expert outputs using downstream data.
}
  \label{fig:palp_overview}
  \vspace{-5pt} 
\end{figure*}

Our proposed framework is deeply motivated by the findings from the preliminary study. It consists of two stages: two-branch pretraining and parameter-efficient adaptation. 
During pretraining, we independently train the node and edge modules on the same dataset to avoid the problem of imbalanced training. To capture the diverse distributions from the pretraining data, an MoE architecture is used to encode different knowledge into distinct experts. During adaptation, we allow different graphs to automatically leverage pretrained knowledge from various experts in a data-driven manner. We detail our design in the following subsections.
The overall framework of PALP is illustrated in Figure~\ref{fig:palp_overview}. 

\subsection{Pretraining with Mixture of Experts}


\textbf{Two-branch pretraining.} 
\label{subsec:pretraining}
Our preliminary study indicates that early fusion hinders effective training of the node module. As a solution, we propose to pretrain the node module and the edge module independently to ensure effective training of both branches. For the node module, we adopt NAGphormer as the backbone. NAGphormer possesses several advantages in the context of LP pretraining: (1) it has a time complexity linear to the number of nodes, enabling large-scale pretraining; (2) it adaptively aggregates information from different localities of a given node, which is favored by link prediction tasks that require personalized receptive fields for different graphs; (3) it adopts a standard Transformer backbone, benefiting from speedup techniques developed specifically for such architecture. Specifically, the node representation given by NAGphormer is:
\begin{equation}
    h_i = \text{NAG}([x_i^0 \mid x_i^1 \mid \dots \mid x_i^K]),
\end{equation}
where $x_i^k$ denotes the aggregated neighborhood around node $i$ at the $k$th hop. To predict the probability of observing an edge $(i, j)$, $h_i$ and $h_j$ are pooled through an element-wise product before being fed into an MLP:
\begin{equation}
    p_{ij} = \sigma(\text{MLP}(h_i \odot h_j)),
\end{equation}
where $\sigma$ denotes the sigmoid function.

For the edge module, we follow our preliminary study and choose the structural embeddings from BUDDY, as given in Equation~\eqref{eq:buddy}. 
\begin{equation}
    \{B_{uv}[d], A_{uv}[d_u, d_v] : \forall d, d_u, d_v \in [k]\}
    \label{eq:buddy}
\end{equation}
where $k$ is the receptive field, $A_{uv}[d_u, d_v]$ denotes the number of nodes at distances exactly $d_u$ and $d_v$ from $u$ and $v$ respectively, and $B_{uv}[d]$ is computed by:
\begin{equation}
B_{uv}[d] = \sum_{d_v=k+1}^{\infty} A_{uv}[d, d_v]
\end{equation}

The node counts $A$ and $B$ from the above equation can be efficiently approximated with the sketching techniques introduced in \cite{Wang2024SubgraphPT}. Such structural features, sometimes referred to as \textit{labeling tricks}, have proven effective to increase the expressiveness of link representation obtained by MPNNs and go beyond the Weisfeiler-Leman (WL) graph isomorphism test to capture critical heuristics for LP \cite{li2020distance, zhang2021labeling}.
We do not use learnable edge modules due to two reasons: (1) any learnable model applied to edges will incur an additional operation of at least $O(E)$, which poses computational challenges for pretraining on large graphs and (2) learnable pairwise embeddings may be too flexible to capture universal LP factors that transfer across datasets. Similar to the node module, we obtain the probabilities of edge existence using an MLP:
\begin{equation}
    p_{ij} = \sigma(\text{MLP}(e_{ij})).
\end{equation}

We use the BinaryCrossEntropy loss to train the two modules. Specifically, 
\begin{equation}
    \mathcal{L} = - \frac{1}{|E^+| + |E^-|} \left( \sum_{(i,j) \in E^+} \log p_{ij} + \sum_{(i,j) \in E^-} \log (1 - p_{ij}) \right)
\end{equation}
where $|E^+|$ and $|E^-|$ denotes the sampled set of positive edges and negative edges, respectively.

\noindent\textbf{MoE architecture.} \
As suggested by previous studies, the benefits from pretraining depend on how much the pretraining distribution matches the downstream data \cite{xu2023betterWITHLESS, cao2023pre}. To expedite transferability while encoding a large diversity of distributions, we propose to model the patterns for link formation with multiple expert models and use a gating function to guide the knowledge flow. Specifically, we adopt an edge-level gating mechanism that routes each input edge to a certain expert, depending on the characteristics of the query edge. The edge-level routing is motivated by the observation that edges with distinct properties tend to benefit from different processing strategies \cite{ma2024mixture}. When pretraining on a large dataset with billions of edges, such flexibility becomes even more critical for encoding the pretraining knowledge in a finer-grained manner. 

For the node module, we use a shared node encoder and adopt multiple score functions as the experts. The encoder acts as a universal feature extractor, while different experts aim to capture distinct patterns for link formation. This design mimics practices from self-supervised learning methods, where one backbone model is trained to work with multiple projectors for different tasks. We empirically found this strategy effective while significantly reducing the model size, as most parameters reside in the node encoder rather than the score functions. Since the edge encoder in BUDDY is non-parametric, we also include multiple score functions to realize MoE in the edge branch. An illustration of the pretraining framework is given in the left part of Figure \ref{fig:palp_overview}.

To avoid the expert collapse problem in training MoE models, we adopt a cluster-based gating mechanism that assigns input edges to experts based on their distances to the corresponding cluster centers, inspired by \cite{kim2023learning}. This approach ensures that each expert models a subset of training data, preventing cases where certain experts were never trained. 
Specifically, given an edge $(i, j)$ and its gating feature $g_{ij}$, we first project $g_{ij}$ to a latent space using an MLP:
\begin{equation}
    z_{ij} = \text{MLP}(g_{ij}).
\end{equation}
Then, we compute the negative Euclidean distances of $z_{ij}$ to all the cluster centers, and use them as the weights for the corresponding experts:
\begin{equation}
    w_{ij}^k = -\lVert z_{ij} - c_k \rVert_2,
\end{equation}
where $c_k$ are learnable clusters in the latent space. 
In our implementation, we represent each edge feature as the sum of its two endpoint node features and use this representation as input to the gating function:
\begin{equation}
    g_{ij} = x_i + x_j.
\end{equation}
This simple sum pooling proves effective due to its discriminative power in determining the relative position of the current edge within the entire pretraining data distribution.

Given weights for different experts, we use Gumbel-Softmax to approximate sampling from the expert distribution while maintaining differentiability of the entire architecture:
\begin{equation}
    p_{ij}^k = \frac{\exp\left((w_{ij}^k + G_k) / \tau\right)}{\sum_{k'} \exp\left((w_{ij}^{k'} + G_{k'}) / \tau\right)},
\end{equation}
where $G_k$ are i.i.d. samples from the $Gumbel(0, 1)$ distribution, and $\tau$ is the temperature parameter. The temperature is adjusted according to the following schedule:
\begin{equation}
    \tau_t = \max\left(\tau_\text{final}, \tau_0 \cdot \alpha^t\right),
\end{equation}
where $\tau_0$ is the initial temperature, $\tau_\text{final}$ is the lower bound, $\alpha$ is the decay rate (a constant less than 1), and $t$ denotes the current training epoch.
In the early stages of training, a high temperature encourages exploration by producing more uniform weights across experts. As training progresses, the temperature gradually decreases, making the gating function more selective and enabling the model to focus on the most relevant experts for each input edge. In our implementation, we set $\alpha = 0.8$, which empirically yields good performance.
In short, the cluster-based expert assignment strategy, combined with temperature annealing, ensures effective load distribution across experts while minimizing knowledge conflicts. 

\subsection{Downstream Adaptation}
\label{subsec:adaptation}
After pretraining on the large-scale dataset, PALP can be used as a zero-shot link predictor, or adapts to downstream datasets through parameter-efficient tuning. 

\noindent{\bf Zero-shot learning.}
In zero-shot setting, we feed the test graph into the node-module and the edge-module independently, and then sum up the sigmoid normalized probablitilies from the two branches as the final predicted scores of link existence. This strategy provides a straightforward way to leverage information from the two branches, while avoiding modifying the pretrained parameters. We denote this approach as \textit{PALP-sum}, and the predicted probability is given by:
\begin{equation}
    p_{ij} = \sigma\left( \sigma(l_{ij}^{N}) + \sigma(l_{ij}^{E}) \right),
\end{equation}
where $l_{ij}^{N}$ and $l_{ij}^{E}$ denote the logits from the node and edge modules, respectively.

\noindent{\bf Parameter-efficient tuning.}
Given the set of pretrained experts, our goal is to find a proper way to combine them that best fits the data at hand. Specifically, different graphs may lean towards the node module or the edge module, or prefer experts trained with specific data distributions. Therefore, a promising solution would allow the downstream data to flexibly select experts from both branches and adaptively merge their outputs.
To achieve this, we propose \textit{PALP-adapt}, which automatically fuses the experts for a given graph:
\begin{equation}
    p_{ij} = \sigma\left( \sum_k p^k \cdot l_{ij}^k \right),
\end{equation}
where $l_{ij}^k$ denotes the logits from the $k$-th expert for edge $(i, j)$, and $p^k$ are the learnable weights, which constitutes a vector $p \in \mathbb{R}^K$. Note that we use the same weight vector $p$ for all edges in a given test graph.
When adapting to new graphs, we keep the pretrained experts frozen and only train the weight vector using binary cross-entropy loss on the training edges. In this way, we allow the downstream dataset to reuse knowledge from a diverse set of link predictors adaptively while maintaining high efficiency.

We use soft aggregation on downstream datasets rather than taking the prediction from the top-1 expert. Since the set of pretrained experts exhibit distinct characteristics and make non-identical predictions, the collaborative use of them may lead to improved performance compared to the best single one, according to the established theory in model ensembles.

\subsection{Complexity Analysis}
\label{subsec:complexity}
We carefully choose the components in PALP to ensure scalability. For link prediction, computation is dominated by two operations: (1) node/edge representation generation via encoders and (2) probability estimation using the score function.

In PALP, BUDDY edge features are generated during preprocessing, incurring no additional computation during training. For node encoding, we adopt NAGphormer, which, after precomputing propagated node features, has a training complexity of $O(NKF^2)$, where $N$ is the number of nodes, $K$ is the number of hops, and $F$ is the feature dimension. For large $N$, mini-batch training can be employed without increasing overall complexity. In contrast, existing GFMs often use subgraph-based methods \cite{liu2023one, li2024zerog, huang2024prodigy}, where a $K$-hop ego-subgraph is extracted for each node to be processed by an MPNN. Such approaches incur repetitive processing of the same nodes in different subgraphs, resulting in a complexity of $O(Nd^KF^2)$, where $d$ is the average degree of the pretraining graph.

For the score function, making predictions for $E$ edges requires $O(EF^2)$ operations. Thus, the overall time complexity of PALP pretraining is $O(NKF^2 + EF^2)$, whereas existing GFMs have a complexity of $O(Nd^KF^2 + EF^2)$.
As a result, PALP achieves significantly higher efficiency in link prediction pretraining, making it feasible for large-scale graphs.





%% file: sections/experiments.tex
\begin{table*}[ht]
\centering
\caption{Performance comparison of zero-shot link prediction. Metric: MRR. }
\resizebox{\textwidth}{!}{%
\begin{tabular}{lccccccc|cccccc}
\toprule
& Cora & Citeseer & Pubmed & Art & Business & Geography & Sociology & Child & History & Photo & Computers & Sportfits & Products \\
\hline              
One4all           & 8.15  & 7.91     & 7.05   & 6.98  & 4.30     & 6.30      & 5.49      & 6.41   & 7.98    & 6.30   & 5.96   & 6.33    & 5.82  \\ 
AnyGraph          & 10.71 & 9.61     & 9.67   & 7.94  & 5.51     & 8.33      & 7.01      & 8.48   & 8.91    & 8.53   & 8.75   & 7.58    & 7.33  \\ 
ZeroG             & 11.86 & 10.75    & 10.32  & 9.06  & 5.82     & 8.92      & 7.25      & 9.11   & 10.02   & 8.41   & 8.65   & 8.15    & 8.93  \\
ZeroG-papers100M  & 36.25 &	34.95    & 48.91  & 36.42 & 19.03    & 34.72     & 30.13     & 18.27  & 32.57   & 11.30  & 12.01  & 16.49   & 19.81 \\
SentenceBert      & 54.96 & 63.28    & 59.27  & 56.75 & 31.14    & 48.89     & 44.11     & 41.38  & 59.60   & 17.76  & 18.89  & 38.78   & 43.34 \\ 
\hline
PALP-sum       & \textbf{70.25} & \textbf{74.70} & \textbf{73.33} & \textbf{66.43} & \textbf{42.15} & \textbf{60.02} & \textbf{58.46} & \textbf{74.04} & \textbf{80.36} & \textbf{67.91} & \textbf{70.20} & \textbf{68.45}   & \textbf{66.67} \\ 
\bottomrule
\end{tabular}}
\label{tab:zero-shot}
\end{table*}

In this section, we conduct extensive experiments to validate the effectiveness and efficiency of PALP. We compare our method against baselines under zero-shot and fine-tuning settings. We also analyze the impact of distribution shifts on downstream performance and examine how key components of PALP contribute to its overall effectiveness. Due to space limitations, additional results and discussions are provided in Appendix \ref{subsec:additional_results}.
\vspace{-2.5mm}
\subsection{Experimental Settings}

\textbf{Datasets.} We pretrain our model on the largest publicly available graph, \textsf{ogbn-papers100M} \cite{hu2020openOGB}, and evaluate it on 13 small-to-medium graphs from two domains.
For in-domain evaluation, we use seven citation networks: Cora, Citeseer, and Pubmed from the Platenoid dataset \cite{yang2016revisiting}, along with Art, Business, Geography, and Sociology from the MAPLE collection \cite{zhang2023effect}. To assess PALP's cross-domain transferability, we include six e-commerce networks processed by \cite{chen2024text}. 
Due to the large size of the original MAPLE and e-commerce datasets, evaluating all baseline models, particularly GNN4LP methods \cite{wang2023neural, yun2021neo} and subgraph-based methods \cite{liu2023one, li2024zerog}, is computationally expensive. To this end, we adopt the \href{http://glaros.dtc.umn.edu/gkhome/views/metis}{METIS} algorithm to partition each dataset into several closely connected components, and use the first partition for evaluation.
Additionally, we incorporate three large-scale datasets to further validate our method's effectiveness and efficiency. The dataset statistics are summarized in Table \ref{tab:dataset_statistics}.

To create an aligned input space between all graphs for pretraining and evaluation, we use SentenceBERT \cite{reimers2019sentence} to generate 384-dimensional text embeddings for each node in the graphs during preprocessing. We also precomputed the structural features using the subgraph sketching technique from BUDDY \cite{Wang2024SubgraphPT} for efficient training. For all datasets, we use the Mean Reciprocal Rank (MRR) with 100 randomly-chosen negative samples as the evaluation metric.

\begin{table*}[ht]
\centering
\caption{Performance comparison of end-to-end training methods. Data split: 40/10/50. Metric: MRR.}
\resizebox{\textwidth}{!}{%
\begin{tabular}{lccccccc|cccccc}
\toprule
 & Cora & Citeseer & Pubmed & Art & Business & Geography & Sociology & Child & History & Photo & Computers & Sportfits & Products \\
\hline
MLP & 54.97 & 59.89 & 66.86 & 56.71 & 40.76 & 55.72 & 48.53 & 70.48 & 64.53 & 56.08 & 59.61 & 66.64 & 72.89 \\
GCN & 53.53 & 61.17 & 70.56 & 62.98 & 41.25 & 55.40 & 48.07 & \textbf{75.79} & 66.41 & \textbf{70.55} & \textbf{70.08} & 68.24 & \textbf{75.60} \\
GraphSAGE & 54.40 & 59.95 & 73.02 & 56.17 & 41.59 & 57.13 & 49.91 & 75.16 & 65.11 & 69.89 & 67.50 & 69.49 & 73.66 \\
\hline
Neo-GNN & 50.52 & 57.23 & 65.68 & 55.87 & 26.40 & 47.63 & 40.34 & 68.59 & 63.89 & 62.67 & 61.94 & 60.14 & 57.79 \\
NCN & 57.47 & 41.44 & 70.46 & 63.43 & 40.90 & 55.92 & 48.62 & 75.23 & 71.26 & 66.55 & 66.79 & 68.69 & 74.38 \\
BUDDY & 58.28 & 63.41 & 69.45 & 63.93 & 39.61 & 56.40 & 50.22 & 72.54 & 67.02 & 66.29 & 68.54 & 69.48 & 74.18\\
LPFormer & 59.52 & 62.18 & \textbf{74.25} & 62.13 & 40.28 & 55.90 & 50.03 & 75.36 & 70.28 & 69.84 & 67.33 & \textbf{70.41} & 73.43 \\
\hline
PALP-adapt & \textbf{63.94} & \textbf{70.73} & 71.33 & \textbf{65.77} & \textbf{42.35} & \textbf{58.96} & \textbf{53.01} & 72.65 & \textbf{74.88} & 63.38 & 63.86 & 64.95 & 69.95 \\
\bottomrule
\end{tabular}}
\label{tab:end2end}
\end{table*}

\noindent\textbf{Zero-shot evaluation.}
We first evaluate the effectiveness of PALP under the zero-shot learning setting. 
For this setting, we adopt recently proposed general graph foundation models as baselines, including one4all \cite{liu2023one}, zeroG \cite{li2024zerog} and anyGraph \cite{xia2024anygraph}. These models were first pretrained using their proposed method and then make predictions on downstream graphs without modifying the model. Due to the scalability issues of these methods indicated in Section~\ref{subsec:complexity}, we were unable to pretrain them all on \textsf{ogbn-papers100M} which was used for pretraining PALP. Instead, we first pretrain these models on \textsf{ogbn-arxiv} and evaluate their zero-shot performance. Based on the results, we select the best-performing baseline, ZeroG, for additional pretraining on \textsf{ogbn-papers100M} to enable a fair comparison with PALP. 
Additionally, we include a simple embedding-based method, which computes link scores using cosine similarity between node embeddings generated by SentenceBERT \cite{reimers2019sentence}.

\noindent\textbf{Fine-tuning evaluation.} For this setting, we compare PALP with two classes of baselines.
The first category encompasses classic models including MLP, GCN \cite{kipf2016semi} and GraphSAGE \cite{hamilton2017inductive}. The second category contains advanced GNN4LP methods, including NCN \cite{wang2023neural}, Neo-GNN \cite{yun2021neo}, BUDDY \cite{Wang2024SubgraphPT} and LPFormer \cite{shomer2024lpformer}. We do not compare with GFMs as they are not designed for the fine-tuning setting. To comprehensively study the performance of methods under different graph sparsity, we conduct experiments on two split ratios, (1) sparse graph evaluation: 40/10/50 of the edges are used for training, validation and test and (2) dense graph evaluation: 80/10/10 of the edges are used for training, validation and test. More experimental details can be found in Appendix~\ref{subsec:additional_results}.




\subsection{Zero-shot Performance Comparison }
 We present the comparison under zero-shot learning in Table \ref{tab:zero-shot}. From the table, PALP outperforms all methods by a significant margin on all datasets. The GFM methods perform poorly on downstream datasets, which can be attributed to two reasons: (1) they are not specifically designed for link prediction, and (2) they were pretrained with limited data, which may not cover the diverse downstream distributions.
 For ZeroG-papers100M, despite using the same pretraining dataset as PALP, its performance falls behind even the simple SentenceBert baseline. This poor performance may partially stem from the constrained computing budget, preventing an extensive hyperparameter search during pretraining, which is unrealistic given its complexity. 
 On the other hand, the textual embeddings from SentenceBert provide a reasonable baseline, indicating that the rich semantics contained in nodes' content can be effectively leveraged for link prediction. However, our PALP still surpasses it, demonstrating effective knowledge transfer from the pretrained  models.

\vspace{-2.0mm}
\subsection{Fine-tuning Performance Comparison}

\textbf{Effectiveness comparison.} We compare fine-tuned PALP, adapted via our parameter-efficient strategy, with traditional training-from-scratch baselines in Table \ref{tab:end2end}. The results show that PALP consistently performs well on citation networks, achieving the best results on 6 out of 7 datasets. For e-commerce data, PALP does not outperform the best baselines on most datasets due to the significant distribution shift between pretraining and downstream datasets. However, it still provides reasonable results for these graphs, especially for the History dataset, where PALP outperforms the second-best method with an MRR of over 3\%. This is due to the high similarity between the History dataset, where node features are descriptions of history books, and the \textsf{ogbn-papers100M} data, which contains literature regarding history researches. Overall, PALP demonstrates strong adaptability to both in-domain and cross-domain graphs, with higher performance on datasets closer to the pretraining distribution. A similar trend is observed with higher training ratios in Table \ref{tab:ft_high}, although PALP's advantage becomes less pronounced. This is expected, as sufficient training data reduces the reliance on knowledge transfer.

\begin{figure}[h]
  \centering
  \includegraphics[scale=0.20]{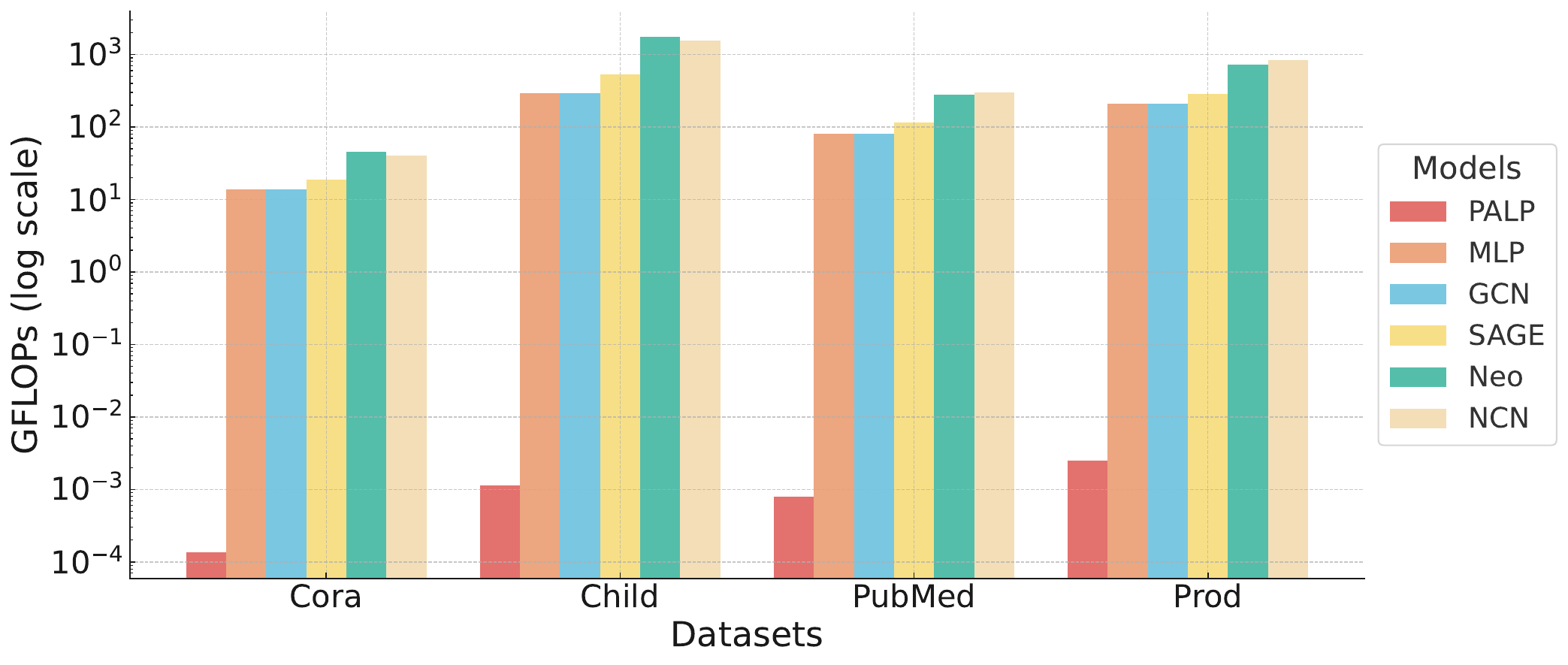} 
  \vspace{-2.5mm}
  \caption{Comparison of per-epoch FLOPs across different methods. Values are shown on a logarithmic scale.}
  \label{fig:flops}
  \vspace{-12pt}
\end{figure}

\noindent\textbf{Efficiency comparision.}
Beyond its strong predictive performance, PALP significantly improves efficiency compared to traditional end-to-end methods. As shown in Figure \ref{fig:flops}, PALP requires over 10,000 times fewer computations per training epoch than baseline models, thanks to its parameter-efficient tuning strategy.
When adapting to unseen graphs, PALP only learns a set of weights to aggregate the output logits from pretrained experts, while keeping all other parameters frozen. This adaptation process involves a single forward pass through the pretrained experts to extract their outputs. Afterward, training reduces to a simple logistic regression with a parameter size equal to the number of experts.
These efficiency gains allow PALP to achieve strong link prediction performance with minimal computational overhead, making it highly practical for real-world applications.

\begin{figure}[h]
  \centering
  \includegraphics[scale=0.25]{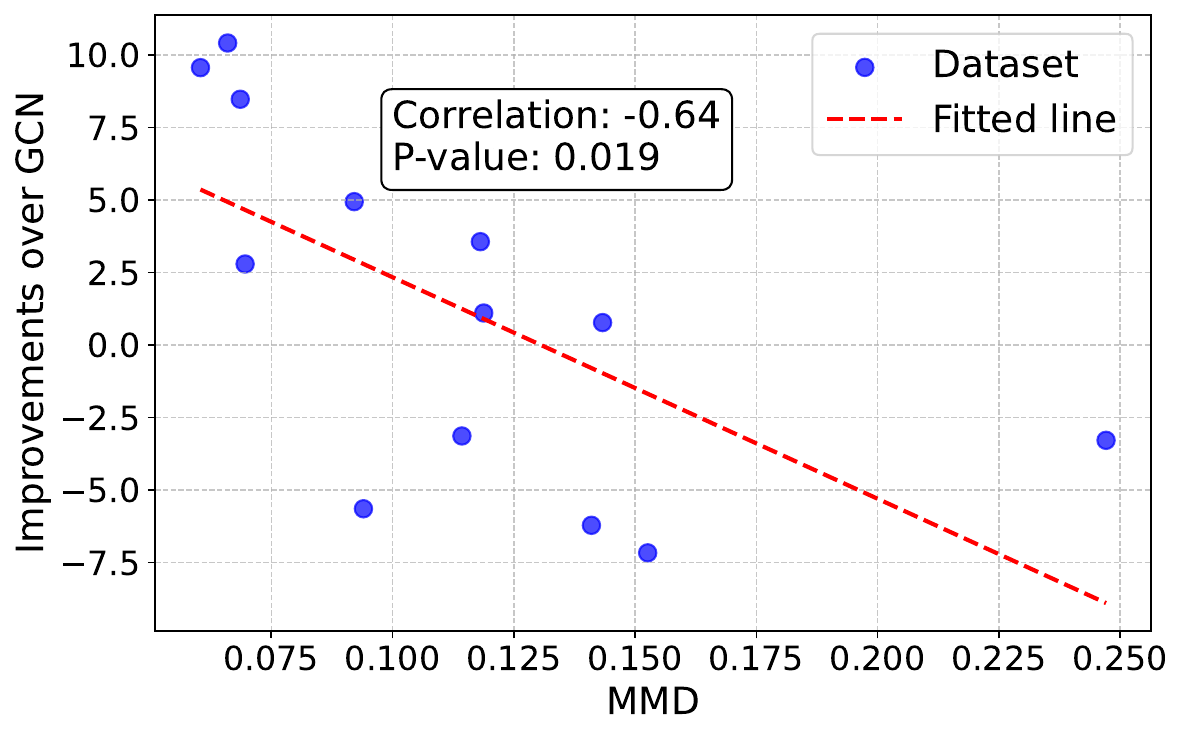} 
  \vspace{-5pt}
  \caption{Correlation between MMD and performance gains.}
  \label{fig:mmd}
  \vspace{-5pt}
\end{figure}

\noindent\textbf{Transferability analysis.}
We investigate the factors influencing PALP's performance across different datasets. Since PALP's parameters remain unchanged during downstream evaluation, its effectiveness is expected to depend on how well the downstream data aligns with the pretraining distribution. To quantify this alignment, we compute the Maximum Mean Discrepancy (MMD) between each downstream dataset and the pretraining data—a widely used metric for measuring distribution shift ~\cite{wang2018visual}.
Figure \ref{fig:mmd} illustrates the correlation between downstream performance and distribution shift, using results from Table \ref{tab:end2end}. The x-axis represents the MMD value, while the y-axis shows PALP's improvement over an end-to-end trained GCN. We observe a negative correlation of -0.64 with a statistical significance of p = 0.019, suggesting that the benefits of pretraining decrease as the distribution shift increases. This suggests the potential of PALP to cover more diverse domains when more pretraining data becomes available. 

\subsection{Ablation Study}
In this section, we evaluate the effectiveness of various components in PALP by introducing the following five variants:

\begin{itemize}
\item \textbf{Node-only}: Utilizes only the output from the node module for link prediction.
\item \textbf{Edge-only}: Utilizes only the output from the edge module for link prediction.
\item \textbf{PALP-sum}: Merges the outputs from the node and edge modules by summing their predicted probabilities.
\item \textbf{PALP-adapt}: Performs parameter-efficient tuning on downstream datasets to learn weights for each expert.
\item \textbf{PALP-adapt w/o MoE}: Similar to PALP-adapt but without the Mixture of Experts (MoE) architecture.
\end{itemize}

\begin{figure}[h]
  \centering
  \includegraphics[scale=0.25]{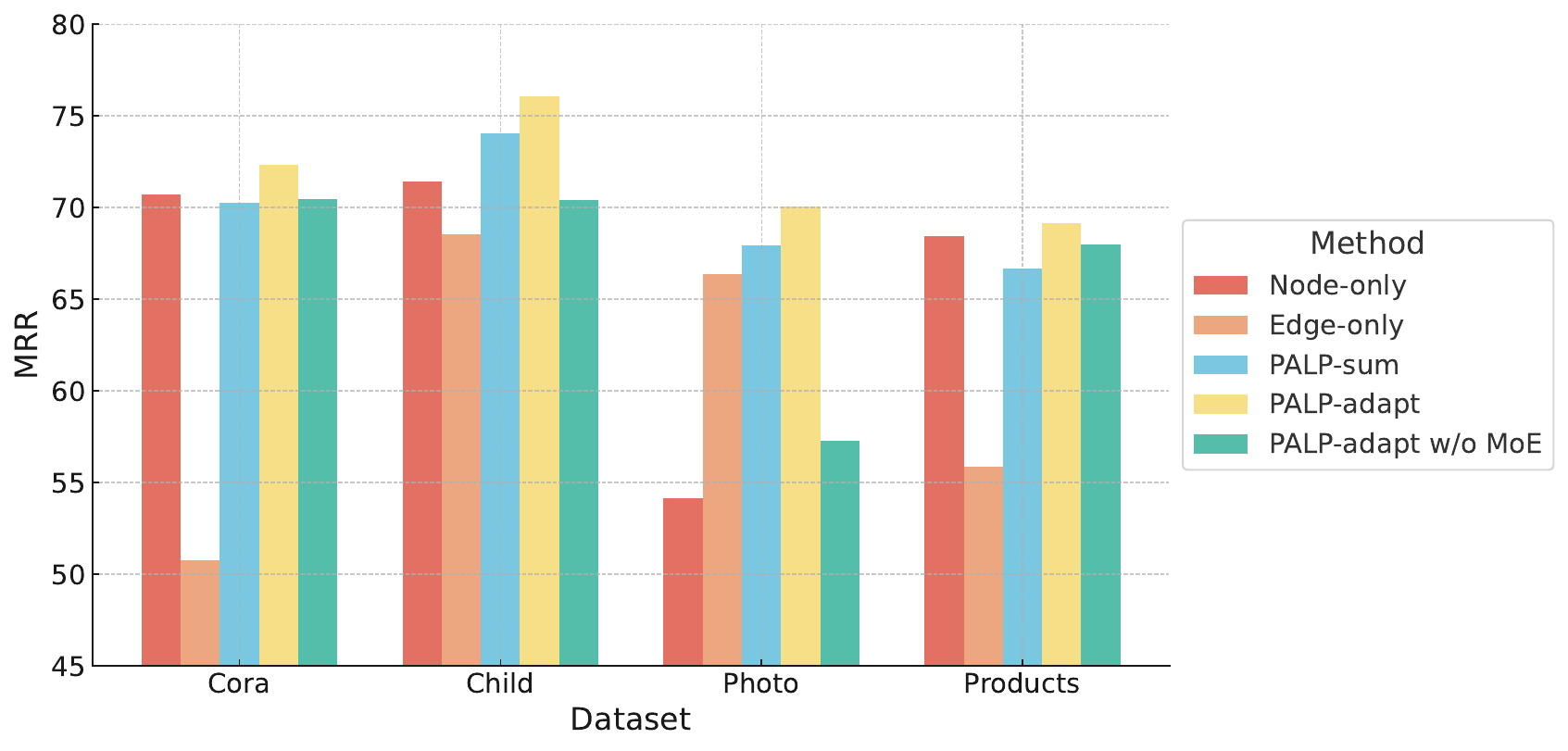} 
   \vspace{-3pt}
  \caption{Ablation study of PALP variants.}
  \label{fig:ablation}
  \vspace{-9pt}
\end{figure}

The results are presented in Figure \ref{fig:ablation}. 
We first observe that the node and edge modules perform optimally on different datasets. For instance, the edge module excels on the Photo dataset, whereas the node module dominates elsewhere. In Child and Photo, summing both modules' outputs improves performance, but in Cora and Products, it degrades performance compared to node-only, suggesting that naïve sum-pooling is insufficient for optimal integration. In contrast, PALP-adapt consistently improves over the individual branches, demonstrating the advantage of adaptively selecting experts based on downstream data. Removing the MoE architecture and using only a single expert per module significantly reduces performance, highlighting the role of MoE in mitigating distribution conflicts. These results confirm that all components of PALP contribute meaningfully to its final performance.

%% file: sections/conclusion.tex

This work introduces a novel pretraining approach for link prediction, leveraging late fusion to effectively combine node- and edge-level information. To enhance transferability across diverse graphs, we propose a mixture-of-experts framework and a parameter-efficient tuning strategy for seamless adaptation with minimal overhead. Our findings establish a foundation for link prediction-specific pretraining, offering a scalable and adaptable solution for real-world graph learning applications. Future work will focus on developing learnable methods to automatically capture various transferable patterns from data, while enhancing out-of-domain transferability.

%% file: sections/appendix.tex


\section{Datasets}
\textbf{Pretraining data.} We use \textsf{ogbn-papers100M} to pretrain PALP. To mitigate out-of-memory issues caused by the large feature matrix and to improve efficiency by avoiding on-the-fly sampling, we preprocess the graph into multiple subgraphs using the METIS algorithm, following \cite{song2024pure}. The statistics of the pretraining dataset are presented in Table \ref{table:partitions}.

\begin{table}[ht]
    \centering
    \caption{Summary of METIS partitions on \textsf{ogbn-papers100M}}  
    \label{table:partitions}
    \resizebox{\columnwidth}{!}{
    \begin{tabular}{c|c|c|c|c}
        \toprule
        \textbf{\#Graphs} & \textbf{Avg. \#Nodes} & \textbf{Avg. \#Edges} & \textbf{\#Node Range} & \textbf{\#Edge Range} \\
        \midrule
        11105 & 10000.90 & 61357.03 & 303 - 45748 & 328 - 122644 \\
        \bottomrule
    \end{tabular}}
\end{table}

\noindent\textbf{Downstream data.} We adopt 16 datasets from two domains for downstream evaluation. Among them, 13 are small to medium-sized, while the remaining 3 are large-scale datasets containing millions of edges. The statistics of the evaluation datasets are presented in Table \ref{tab:dataset_statistics}.

\begin{table}[h!]
\centering
\caption{Statistics of downstream datasets}
\begin{tabular}{lccl}
\hline
\textbf{Dataset Name} & \textbf{\#Nodes} & \textbf{\#Edges} & \textbf{Domain} \\ \hline
Cora & 2,708 & 10,858 & Citation \\ 
Citeseer & 3,186 & 8,554 & Citation \\ 
Pubmed & 19,717 & 88,670 & Citation \\ 
Art & 58,373 & 7,184 & Citation \\
Business & 4,279 & 36,697 & Citation \\
Geography & 7,395 & 32,818 & Citation \\
Sociology & 4,518 & 14,801 & Citation \\
History & 4,153 & 12,622 & E-commerce \\ 
Child & 3,819 & 45,408 & E-commerce \\ 
Photo & 4,865 & 39,081 & E-commerce \\ 
Computers & 4,369 & 33,493 & E-commerce \\ 
Sportsfit & 3,508 & 22,220 & E-commerce \\ 
Products & 3,081 & 196,115 & E-commerce \\ 
CSRankings & 263,393 & 1,464,679 & Citation \\ 
Economics & 178,670 & 1,532,072 & Citation \\ 
Computer Science & 410,603 & 1,494,272 & Citation \\ \hline
\end{tabular}
\label{tab:dataset_statistics}
\end{table}

\section{Experiments}
\subsection{Hyperparameter Settings}
\label{subsec:hyperparams}

\textbf{Hyperparameter selection.} The hyperparameters for the pretraining stage were primarily determined through empirical evaluation. To ensure generalization across various downstream datasets, we monitored two key criteria: (1) training loss and (2) average downstream performance. Table~\ref{tab:hyperparams} summarizes the hyperparameter configurations used in pretraining.

\begin{table}[h]
\centering
\caption{Hyperparameter configurations used in pretraining.}
\label{tab:hyperparams}
\resizebox{\columnwidth}{!}{%
\begin{tabular}{lccccccccc}
\toprule
\textbf{Name} & \textbf{peak\_lr} & \textbf{end\_lr} & \textbf{warmup} & \textbf{epochs} & \textbf{hops} & \textbf{experts} & \textbf{dropout} & \textbf{hidden\_dim} & \textbf{layers} \\
\midrule
\textbf{Value} & $1\text{e-}4$ & $1\text{e-}5$ & $10,000$ & $10$ & $3$ & $4$ & $0.1$ & $768$ & $2$ \\
\bottomrule
\end{tabular}%
}
\end{table}

Among the hyperparameters, the number of experts, number of hops, and hidden dimension size had the most significant impact on performance. Tables~\ref{tab:hidden_dim}, \ref{tab:nag_hops}, and \ref{tab:experts} present a comparative analysis of these factors across different datasets.

\begin{table}[h]
\centering
\caption{Performance comparison with different hidden dimensions.}
\label{tab:hidden_dim}
\begin{tabular}{lcccc}
\toprule
\textbf{hidden\_dim} & \textbf{Cora} & \textbf{Citeseer} & \textbf{Photo} & \textbf{Computer} \\
\midrule
384 & 0.6837 & 0.7214 & 0.6779 & 0.6851 \\
768 & 0.7063 & 0.7483 & 0.6894 & 0.6937 \\
\bottomrule
\end{tabular}
\end{table}

\begin{table}[h]
\centering
\caption{Impact of the number of hops in NAGphormer on performance.}
\label{tab:nag_hops}
\begin{tabular}{lcccc}
\toprule
\textbf{\#NAG hops} & \textbf{Cora} & \textbf{Citeseer} & \textbf{Photo} & \textbf{Computer} \\
\midrule
NAG-2-hop & 0.7069 & 0.7458 & 0.6433 & 0.6515 \\
NAG-3-hop & 0.7063 & 0.7483 & 0.6894 & 0.6937 \\
NAG-6-hop & 0.6889 & 0.7159 & 0.6742 & 0.6825 \\
\bottomrule
\end{tabular}
\end{table}

\begin{table}[h]
\centering
\caption{Performance comparison with different numbers of experts in MoE.}
\label{tab:experts}
\begin{tabular}{lcccc}
\toprule
\textbf{Experts} & \textbf{Cora} & \textbf{Child} & \textbf{Photo} & \textbf{Products} \\
\midrule
w/o MoE & 70.45 & 70.42 & 57.30 & 67.99 \\
2 experts & 71.99 & 75.97 & 66.35 & 69.23 \\
4 experts & 72.35 & 76.07 & 70.04 & 69.17 \\
8 experts & 72.05 & 76.12 & 69.67 & 68.14 \\
\bottomrule
\end{tabular}
\end{table}

\noindent\textbf{Adapting PALP.}
We adopt gradient descent to optimize the weights for expert assignment. An Adam optimizer with a learning rate of 0.001 is used across all experiments. 

\noindent\textbf{Baselines.}
We carefully tune the hyperparameters, including learning rate, weight decay, and number of model layers for MLP, GCN and GraphSAGE using the validation set. 
For GNN4LP methods, we adopt the hyperparameters reported in \cite{li2024evaluating} and carefully tune the learning rate to avoid bad results.

\subsection{Additional Results}
\label{subsec:additional_results}
\textbf{Dense graph evaluation.} 
We assess the performance of fine-tuned PALP under a high training ratio using an 80/10/10 edge split in Table~\ref{tab:ft_high}. The results exhibit similar patterns to those observed with lower training ratios, as shown in Table \ref{tab:end2end}, where PALP demonstrates superior performance on datasets that align closely with the pretraining distribution.

\begin{table*}[ht]
  \caption{Performance comparison of end-to-end training methods. Data split: 80/10/10. Metric: MRR.}
  \resizebox{\textwidth}{!}{%
  \begin{tabular}{lccccccc|ccccccc}
    \toprule
     & Cora & Citeseer & Pubmed & Art & Business & Geography & Sociology & Child & History & Photo & Computers & Sportsfits & Products \\
    \hline
    MLP    & 63.23 & 67.59   & 74.30  & 61.47 & 44.28   & 58.96    & 55.47     & 74.21     & 74.21   & 70.57 & 70.32 & 72.92  & 77.35 \\
    GCN    & 67.97 & 66.68   & 79.44  & 66.52 & 46.54   & 60.48    & 56.16     & 81.09     & 77.10   & \textbf{78.13} & 76.79 & \textbf{75.87}  & 77.57 \\
    GraphSAGE   & 65.00 & 68.19   & \textbf{79.99}  & 62.88 & 45.97   & 61.46    & 58.00     & 80.10     & 76.71   & 76.47 & 74.38 & 74.62  & 77.18 \\
    \hline
    Neo-GNN    & 62.34 & 64.53   & 70.37  & 60.85 & 29.51   & 50.67    & 48.51     & 65.01     & 72.34   & 68.82 & 69.93 & 64.76  & 35.07 \\
    NCN    & 67.52 & 73.54   & 76.27  & \textbf{67.59} & \textbf{46.83}   & 62.69   & 57.05     & 81.72     & 80.12   & 76.47 & 75.36 & 75.25  & \textbf{78.71} \\
    BUDDY &  59.16 & 64.91   & 71.53  & 65.43 & 41.07   & 58.23    & 52.19     & 74.16     & 69.54   & 67.74 & 70.22 & 71.76 & 75.88 \\
    LPFormer & 69.12 & 72.46 & 79.32 & 67.63 & 46.93 & 61.88 & 59.03 & \textbf{82.29} & 79.73 & 77.45 & \textbf{77.78} & 75.15 & 76.03 \\
    \hline
    PALP-adapt   & \textbf{72.35} & \textbf{75.97}   & 74.68  & 66.33 & 44.53   & \textbf{63.07}    & \textbf{60.06}     & 76.07     & \textbf{80.97}   & 70.05 & 71.05 & 71.59  & 69.17 \\
    \bottomrule
  \end{tabular}}
  \label{tab:ft_high}
\end{table*}

\noindent\textbf{Large graph evaluation.} 
We evaluate the performance of PALP on three large-scale graphs, comparing it with two baseline methods in terms of link prediction accuracy and training costs. We use a 1:1:1 split for training, validation, and test edges.
For PALP, the reported running time consists of two parts: an initial forward pass to compute logits from all experts (performed only once) and the subsequent training time per epoch. In contrast, the time reported for baseline methods corresponds to the training duration per epoch. 
The results show that PALP consistently outperforms baseline methods in both predictive performance and computational efficiency.

\begin{table}[ht]
    \centering
    \caption{Performance and runtime comparison on large-scale datasets. MRR is reported for performance evaluation, and time per epoch is measured in seconds. For PALP-adapt, the first number represents the forward pass (performed once), and the second represents training time per epoch.}
    \label{tab:large_graph}
    \resizebox{0.9\columnwidth}{!}{ 
    \begin{tabular}{lccc}
        \toprule
        Method & CSRankings & Economics & Computer Science \\
        \midrule
        \multicolumn{4}{c}{\textbf{MRR (↑)}} \\
        \midrule
        GCN & 0.8032 & 0.7382 & 0.8151 \\
        BUDDY & 0.7998 & 0.7405 & 0.8072 \\
        PALP-adapt & \textbf{0.8181} & \textbf{0.7537} & \textbf{0.8286} \\
        \midrule
        \multicolumn{4}{c}{\textbf{Training Time (↓)}} \\
        \midrule
        GCN & 139.35s & 111.41s & 237.35s \\
        BUDDY & 160.42s & 120.19s & 289.52s \\
        PALP-adapt & \textbf{4.10s + 0.56s} & \textbf{3.24s + 0.58s} & \textbf{4.52s + 1.54s} \\
        \bottomrule
    \end{tabular}
    }
\end{table}

\begin{figure}[h]
  \centering
  \includegraphics[width=\linewidth]{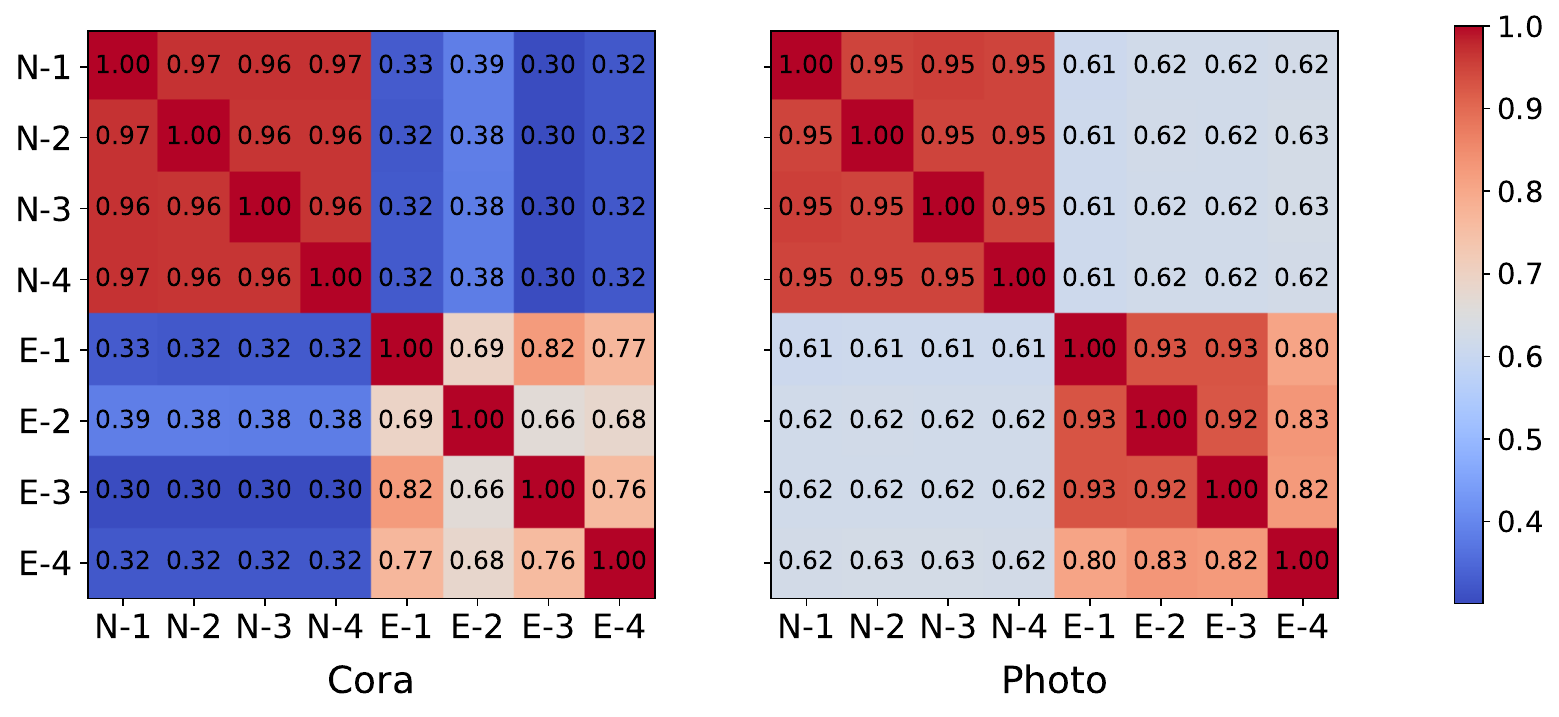} 
  \caption{Similarity between pretrained experts. Values indicate the overlap in correct predictions between pairs of experts. Results are based on Cora and Photo datasets.
    }
  \label{fig:heatmap}
\end{figure}

\noindent\textbf{Expert analysis.}
We analyze the distinctiveness of different experts obtained during pretraining to understand whether they behave similarly or differently on downstream datasets. Given PALP’s dual-branch design, it consists of 
$m$ node experts and $n$ edge experts, each sharing the same architecture but trained on different subsets of the pretraining data. To quantify their differences, we follow \cite{ma2024mixture} and compute the Jaccard similarity between pairs of experts based on their correctly predicted edges, where an edge is considered correctly predicted if the ground-truth target node is ranked among the top 3 out of 100 randomly sampled nodes. Figure \ref{fig:heatmap} presents the overlap ratio between 8 experts on Cora and Photo, where N denotes node experts and E denotes edge experts. We observe that (1) experts trained on the same type of information tend to exhibit similar but not identical behavior, and (2) experts trained on different input information behave distinctly, suggesting that node and edge experts focus on complementary aspects of link prediction. These findings confirm that different experts specialize in distinct predictive patterns, reinforcing the potential for effective model fusion in PALP.

